\documentclass[pdflatex,sn-mathphys-num]{sn-jnl}% Math and Physical Sciences Numbered 

\usepackage{graphicx}%
\usepackage{multirow}%
\usepackage{amsmath,amssymb,amsfonts}%
\usepackage{amsthm}%
\usepackage{mathrsfs}%
\usepackage[title]{appendix}%
\usepackage{xcolor}%
\usepackage{textcomp}%
\usepackage{manyfoot}%
\usepackage{booktabs}%
\usepackage{algorithm}%
\usepackage{algorithmicx}%
\usepackage{algpseudocode}%
\usepackage{listings}%

\usepackage{makecell}
\usepackage{silence}
\usepackage[switch]{lineno}
\usepackage[utf8]{inputenc}
\usepackage{array}
\usepackage{ulem}

\usepackage{amsmath, amssymb, amsthm, physics}
\usepackage{booktabs, multirow}

\newcommand{\vect}[1]{\mathbf{#1}}

\usepackage{bm}

\begin{document}

% \linenumbers

\title[Article Title]{PA-SFM: Tracker-free differentiable acoustic radiation for freehand 3D photoacoustic imaging}

\author[1]{\fnm{Shuang} \sur{Li}}
\equalcont{These authors contributed equally to this work.}

\author[2]{\fnm{Jian} \sur{Gao}}
\equalcont{These authors contributed equally to this work.}

\author[3]{\fnm{Chulhong} \sur{Kim}}

\author[3]{\fnm{Seongwook} \sur{Choi}}

\author[1]{\fnm{Qian} \sur{Chen}}

\author[1]{\fnm{Yibing} \sur{Wang}}

\author[2]{\fnm{Shuang} \sur{Wu}}

\author[1]{\fnm{Yu} \sur{Zhang}}

\author[1]{\fnm{Tingting} \sur{Huang}}

\author[1]{\fnm{Yucheng} \sur{Zhou}}

\author[1]{\fnm{Boxin} \sur{Yao}}

\author*[2]{\fnm{Yao} \sur{Yao}}\email{yaoyao@nju.edu.cn}

\author*[1, 4]{\fnm{Changhui} \sur{Li}}\email{chli@pku.edu.cn}

\affil*[1]{\orgdiv{Department of Biomedical Engineering, College of Future Technology}, \orgname{Peking University}, \city{Beijing}, \country{China}}

\affil*[2]{\orgdiv{School of Intelligence Science and Technology}, \orgname{Nanjing University}, \city{Suzhou}, \country{China}}

\affil[3]{\orgdiv{Department of Electrical Engineering, Convergence IT Engineering, Mechanical Engineering, and Medical Science and Engineering, Medical Device Innovation Center}, \orgname{Pohang University of Science and Technology}, \city{Pohang}, \country{Republic of Korea}}

\affil*[4]{\orgdiv{National Biomedical Imaging Center}, \orgname{Peking University}, \city{Beijing}, \country{China}}

\abstract{Three-dimensional (3D) handheld photoacoustic tomography typically relies on bulky and expensive external positioning sensors to correct motion artifacts, which severely limits its clinical flexibility and accessibility. To address this challenge, we present PA-SFM, a tracker-free framework that leverages exclusively single-modality photoacoustic data for both sensor pose recovery and high-fidelity 3D reconstruction via differentiable acoustic radiation modeling. Unlike traditional structure-from-motion (SFM) methods based on visual features, PA-SFM integrates the acoustic wave equation into a differentiable programming pipeline. By leveraging a high-performance, GPU-accelerated acoustic radiation kernel, the framework simultaneously optimizes the 3D photoacoustic source distribution and the sensor array pose via gradient descent. To ensure robust convergence in freehand scenarios, we introduce a coarse-to-fine optimization strategy that incorporates geometric consistency checks and rigid-body constraints to eliminate motion outliers. We validated the proposed method through both numerical simulations and \textit{in-vivo} rat experiments. The results demonstrate that PA-SFM achieves sub-millimeter positioning accuracy and restores high-resolution 3D vascular structures comparable to ground-truth benchmarks, offering a low-cost, software-defined solution for clinical freehand photoacoustic imaging. The source code is publicly available at \href{https://github.com/JaegerCQ/PA-SFM}{https://github.com/JaegerCQ/PA-SFM}.}


\maketitle

\section{Introduction}\label{sec1}

Photoacoustic imaging (PAI) is a hybrid imaging modality that leverages ultrasound detection alongside optical absorption contrast, enabling non-invasive visualization of biological tissues at centimeter-scale depths with spatial resolutions superior to conventional optical imaging techniques. Due to these advantages, PAI has found wide application in both preclinical studies and clinical settings~\cite{park2025clinical, wang2012photoacoustic, assi2023review, dean2017advanced, lin2022emerging, ntziachristos2025addressing}. Advancements in three-dimensional (3D) PAI using 2D matrix arrays have enabled \textit{in-vivo} imaging of peripheral vessels~\cite{wray2019photoacoustic, li2024photoacoustic}, breast tissues~\cite{wang2021functional, han2021three}, and small animals~\cite{sun2024real} with promising results. Various array designs, including spherical, planar~\cite{matsumoto2018label, matsumoto2018visualising, ivankovic2019real, dean2013portable, nagae2018real, kim2023wide, piras2009photoacoustic, heijblom2012visualizing}, and synthetically scanned arrays~\cite{li2024photoacoustic, wang2024comprehensive}, have been explored to enhance 3D imaging capabilities.

Despite the diagnostic advantages of 3D photoacoustic tomography (PACT), its integration into routine clinical practice is often hindered by hardware constraints. Conventional high-fidelity 3D reconstruction typically necessitates the use of large-scale 2D matrix arrays or specialized motorized scanning stages to ensure precise spatial registration of captured data~\cite{wang2025cross, zhang2026rotational, li2024photoacoustic, park2025clinical}. While these systems provide controlled trajectories, they are often bulky, expensive, and lack the mechanical flexibility required to navigate irregular anatomical surfaces or operate in resource-limited point-of-care settings~\cite{lee2025enhancing}. Handheld, freehand scanning has emerged as a compelling alternative, offering clinicians unrestricted flexibility to cover large regions of interest. However, the absence of a fixed mechanical reference introduces significant out-of-plane motion artifacts. To achieve high-fidelity volumetric reconstruction, the sensor's six-degree-of-freedom (6-DoF) pose must be accurately recovered for every frame~\cite{lee2025enhancing}. Failure to account for these pose variations leads to severe structural blurring, vessel fragmentation, and loss of quantitative accuracy.

To compensate for manual motion, traditional solutions rely on external positioning sensors such as optical tracking systems (OTS), electromagnetic (EM) trackers, GPS sensors or inertial measurement units (IMUs)~\cite{dou2024sensorless, lee2023review}. Although OTS can provide sub-millimeter precision, it requires a continuous line-of-sight between the camera and the probe markers, which is often impractical in crowded clinical environments or surgical suites~\cite{dou2024sensorless, lasso2014plus}. EM tracking avoids line-of-sight issues but is highly susceptible to electromagnetic interference from metallic medical tools or nearby electronic equipment~\cite{lee2023review}. IMU-based solutions are cost-effective and portable but suffer from inherent cumulative drift, where small measurement errors integrate over time to cause significant positioning offsets~\cite{dou2024sensorless, frey2025ultraflex}. These hardware-based methods increase system complexity, necessitate tedious calibration procedures, and significantly raise the overall cost of the imaging platform.

In response, recent research has shifted toward tracker-free or sensorless pose estimation directly from acquired data. In ultrasound imaging, speckle tracking is a well-established technique for motion estimation; however, photoacoustic signals are inherently speckle-free. Because PAI signals originate from the thermoelastic expansion of discrete absorbers, all initial pressure rises are positive, lacking the phase-randomized interference patterns that create ultrasound speckle. Consequently, traditional cross-correlation algorithms often fail in the photoacoustic domain. Emerging deep learning methods have attempted to learn motion priors from 2D images, yet these models typically require massive, high-quality labeled datasets that are difficult to obtain and fail to transfer across different imaging systems, which lack a rigorous connection to the underlying physics of acoustic wave propagation~\cite{lee2025enhancing}. Furthermore, some researchers have utilized the frequency-domain characteristics of vascular networks in photoacoustic images to estimate the relative motion and orientation of the imaging probe~\cite{knauer2019spatial}, thereby enabling freehand imaging. Although this approach expands the field of view to some extent, significant limitations remain. Because the estimation relies on images rather than the intrinsic properties of the photoacoustic equation, this method strictly requires a high overlap ratio between different poses. Translation estimation may fail if the displacement is excessively large or if the images themselves lack distinct features. Additionally, the scanning direction cannot change abruptly, and the moving speed must be restricted.

Inspired by the success of differentiable rendering in computer vision~\cite{schonberger2016structure}, we present PA-SFM, a tracker-free framework that leverages exclusively single-modality photoacoustic data for both sensor pose recovery and high-fidelity 3D reconstruction. Unlike traditional structure-from-motion methods based on visual features, PA-SFM integrates the acoustic wave equation into a differentiable programming pipeline. By representing the initial pressure field as a superposition of continuous Gaussian kernels, we utilize a GPU-accelerated acoustic radiation kernel to simultaneously optimize the 3D source distribution and the sensor array pose via gradient descent~\cite{tillet2019triton}. To ensure robust convergence in freehand scenarios, we introduce a coarse-to-fine optimization strategy incorporating geometric consistency checks and rigid-body constraints to eliminate motion outliers. We validated the proposed method through numerical simulations and \textit{in-vivo} rat experiments, demonstrating sub-millimeter positioning accuracy. Furthermore, by utilizing a sparse planar array at multiple poses to synthesize an equivalent array with an ultra-wide view with PA-SFM, we achieved high-quality three-dimensional reconstruction. PA-SFM offers a low-cost, software-defined solution that significantly improves the clinical flexibility and accessibility of freehand 3D photoacoustic imaging.

\section{Methods}\label{sec2}

\begin{figure}[htbp]
\centering
\includegraphics[width=1.0\textwidth]{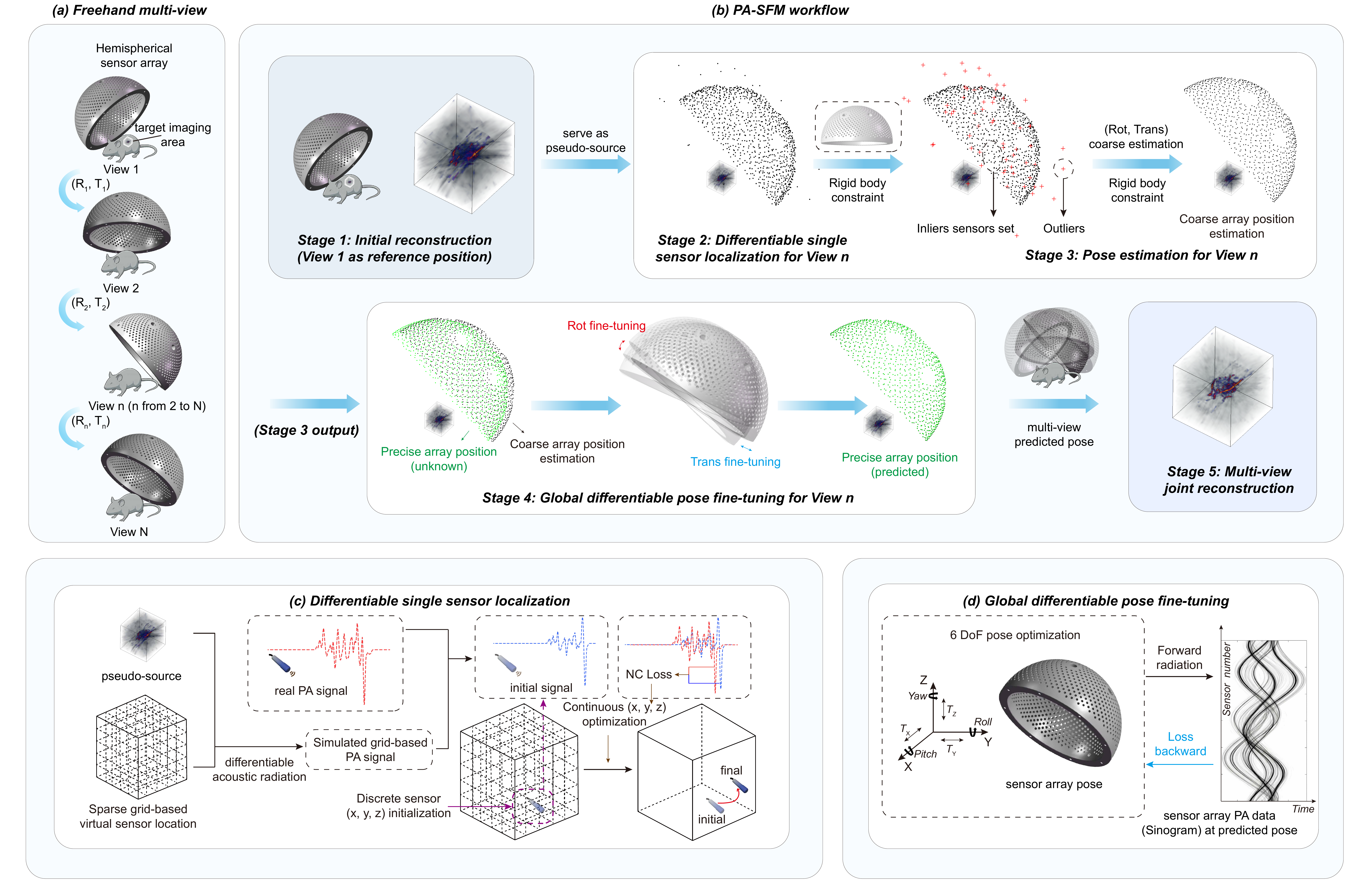}
\caption{The overview framework of PA-SFM algorithm for freehand multi-view 3D PAI. (a) The freehand multi-view 3D PAI. (b) The PA-SFM pipeline. (c) The differentiable single probe localization of PA-SFM. (d) The global differentiable pose fine-tuning of PA-SFM.}\label{methods}
\end{figure}

The proposed PA-SFM framework establishes a complete closed-loop pipeline from reference reconstruction to unknown pose estimation, culminating in joint multi-view reconstruction (Fig.~\ref{methods}). The system is formulated upon a custom-built differentiable acoustic radiation engine and constrained by rigid-body geometry. The detailed algorithmic workflow is structured into the following interacting modules.

\subsection*{Differentiable Acoustic Radiation}
At the core of the framework is a high-performance differentiable acoustic Radiation engine that models forward acoustic wave propagation and computes gradients. To map the 3D photoacoustic initial pressure distribution $\mathbf{P}_c \in \mathbb{R}^{N_x \times N_y \times N_z}$ to the time-resolved sensor signals $p(\mathbf{x}_s, t)$ at sensor location $\mathbf{x}_s \in \mathbb{R}^3$, we employ a Gaussian kernel-based propagation model. The forward model is defined as:
\begin{equation}
    p(\mathbf{x}_s, t) = \sum_{\mathbf{x} \in \Omega} \mathbf{P}_c(\mathbf{x}) \cdot \mathcal{G}(\|\mathbf{x} - \mathbf{x}_s\|, t; \sigma)
\end{equation}
where $\Omega$ denotes the reconstructed volume, and $\mathcal{G}$ is the spatio-temporal Gaussian kernel parameterized by width $\sigma$. 

The physical validity and mathematical rigor of this Gaussian kernel-based discrete model are rooted in the fundamental photoacoustic wave equation under far-field approximations. A comprehensive analytical derivation, which starts from the generalized spherically symmetric initial conditions and culminating in the exact GPU-accelerated far-field discrete model used in our engine, is provided in \textbf{Appendix \ref{appendix:theoretical_derivation}}.

To achieve scalable and highly concurrent GPU acceleration, the forward and backward propagation operators are implemented as custom CUDA kernels using OpenAI Triton. By encapsulating these kernels within \texttt{torch.autograd.Function}, the engine seamlessly supports bidirectional gradient flow with respect to both the source intensity $\mathbf{P}_c$ and the sensor spatial coordinates $\mathbf{x}_s$. This automatic differentiation capability provides a unified mathematical foundation for both image reconstruction and sensor localization.

\subsection*{Stage 1: Initial Reference Reconstruction}
The first stage establishes a structural reference map (coarse model) using data acquired at a known, fixed pose (Pose A). Given the known sensor array coordinates $\mathbf{X}_A$ and the corresponding measured signals $\mathbf{S}_A$, the initial source distribution $\mathbf{P}_c$ is initialized on a $256^3$ voxel grid. The reconstruction is formulated as an inverse optimization problem:
\begin{equation}
    \hat{\mathbf{P}}_c = \arg\min_{\mathbf{P}_c} \left( \| \mathcal{F}(\mathbf{P}_c, \mathbf{X}_A) - \mathbf{S}_A \|_2^2 + \lambda \mathcal{L}_{\text{TGV}}(\mathbf{P}_c) \right)
\end{equation}
where $\mathcal{F}$ represents the forward simulation engine, and $\mathcal{L}_{\text{TGV}}$ is the Total Generalized Variation (TGV) regularization term utilized to preserve structural edges while suppressing background noise. The objective function is iteratively minimized using the Adam optimizer, yielding the initial 3D anatomical structure serving as the reference map.

\subsection*{Stage 2: Differentiable Single Sensor Localization}
To estimate the unknown sensor pose (Pose B), we independently localize each transducer element using the reference map $\hat{\mathbf{P}}_c$ obtained in Stage 1. By fixing $\hat{\mathbf{P}}_c$, the individual sensor coordinates $\mathbf{x}_{s,i}$ become the optimizable parameters. The localization is driven by a Negative Correlation (NC) loss:
\begin{equation}
    \mathcal{L}_{\text{NC}}(\mathbf{x}_{s,i}) = - \frac{\text{Cov}(\mathcal{F}(\hat{\mathbf{P}}_c, \mathbf{x}_{s,i}), \mathbf{S}_{B,i})}{\sigma_{\text{sim}} \sigma_{\text{meas}}}
\end{equation}

To circumvent local minima, a coarse-to-fine search strategy is employed, as demonstrated in Fig.~\ref{methods}(c):
\begin{enumerate}
    \item \textbf{Global Coarse Search:} The target space is discretized into grids, and candidate locations are ranked based on $\mathcal{L}_{\text{NC}}$ to select the Top-$K$ priors.
    \item \textbf{Gradient-based Refinement:} Starting from the candidate priors, gradient descent is utilized to optimize the $(x, y, z)$ coordinates. 
    \item \textbf{Dynamic Smoothing:} The Gaussian kernel width $\sigma$ in Eq. (1) is dynamically decayed (from large to small) during optimization, acting as a simulated annealing mechanism to ensure convergence to the global optimum.
\end{enumerate}
This stage outputs a set of independently predicted, albeit noisy, sensor coordinates $\tilde{\mathbf{X}}_B$.

\subsection*{Stage 3: Robust Rigid Array Estimation}
To eliminate the localization noise introduced in Stage 2, the inherent geometric constraints of the rigid sensor array are leveraged. Given the predefined geometric template of the array $\mathbf{X}_{\text{template}}$, the problem is transformed into finding the optimal global rotation matrix $\mathbf{R} \in SO(3)$ and translation vector $\mathbf{t} \in \mathbb{R}^3$. 

A modified Random Sample Consensus (RANSAC) algorithm is implemented. To drastically improve search efficiency under low inlier ratios, a \textit{geometric consistency pre-check} is introduced: randomly sampled point sets are evaluated based on their pairwise edge distances before performing Singular Value Decomposition (SVD). For valid subsets, the Kabsch algorithm is applied to solve the rigid transformation. This stage yields a corrected global array coordinate set $\mathbf{X}_B^{\text{corr}}$ and reliably identifies a subset of inlier sensors $\mathcal{I}$.

\subsection*{Stage 4: Global Differentiable Fine-tuning}
To further enhance spatial precision, a \texttt{RigidArrayOptimizer} module is constructed. The optimization variables are globally parameterized as Euler angles $\bm{\theta}$ and the translation vector $\mathbf{t}$ (Fig.~\ref{methods}(d)), ensuring that the array strictly adheres to rigid body kinematics: $\mathbf{X}_B(\bm{\theta}, \mathbf{t}) = \mathbf{R}(\bm{\theta})\mathbf{X}_{\text{template}} + \mathbf{t}$.

We introduce an inlier-masked gradient flow mechanism. The global loss is computed exclusively on the highly reliable signals from the inlier set $\mathcal{I}$ (identified in Stage 3):
\begin{equation}
    \mathcal{L}_{\text{global}}(\bm{\theta}, \mathbf{t}) = \sum_{i \in \mathcal{I}} \mathcal{L}_{\text{NC}} \left( \mathbf{R}(\bm{\theta})\mathbf{x}_{\text{template},i} + \mathbf{t} \right)
\end{equation}
Crucially, while the loss is masked, the gradients backpropagate to update the global parameters $(\bm{\theta}, \mathbf{t})$. This mechanism physically compels the unreliable (outlier) sensors to be guided into their correct anatomical positions by the reliable (inlier) sensors, yielding highly precise Pose B coordinates.

\subsection*{Stage 5: Joint Multi-view Reconstruction}
In the final stage, the data from the known Pose A ($\mathbf{X}_A, \mathbf{S}_A$) and the calibrated Pose B ($\mathbf{X}_B^{\text{fine}}, \mathbf{S}_B$) are seamlessly fused. The objective function defined in Stage 1 is re-executed utilizing the augmented synthetic aperture. By coherently combining the dual-view data, limited-view artifacts are effectively mitigated, resulting in the final high-resolution, artifact-free 3D photoacoustic volume.

\begin{algorithm}[htbp]
\caption{Differentiable Photoacoustic Structure from Motion (PA-SFM)}
\label{alg:pa_sfm}
\begin{algorithmic}[1] % [1] enables line numbers

\Require
    Known Pose A data: $\mathbf{X}_A$, $\mathbf{S}_A$; 
    Unknown Pose B signals: $\mathbf{S}_B$;
    Rigid array template: $\mathbf{X}_{\text{template}}$;
    Simulation engine $\mathcal{F}$ with TGV weight $\lambda$.
\Ensure 
    High-resolution joint 3D volume $\mathbf{P}_{\text{final}}$; 
    Fine-tuned Pose B array coordinates $\mathbf{X}_B^{\text{fine}}$.
\Statex
\State \textbf{ Stage 1: Initial Reference Reconstruction}
\State Initialize source distribution $\mathbf{P}_c$ on a 3D grid.
\While{not converged}
    \State $\hat{\mathbf{P}}_c \leftarrow \arg\min_{\mathbf{P}_c} \left( \| \mathcal{F}(\mathbf{P}_c, \mathbf{X}_A) - \mathbf{S}_A \|_2^2 + \lambda \mathcal{L}_{\text{TGV}}(\mathbf{P}_c) \right)$ 
\EndWhile

\Statex
\State \textbf{ Stage 2: Differentiable Single Sensor Localization}
\State Initialize empty noisy coordinate set $\tilde{\mathbf{X}}_B \leftarrow \emptyset$
\For{each sensor $i = 1, 2, \dots, N$ in Pose B}
    \State $\mathbf{x}_{s,i}^{0} \leftarrow$ Top-$1$ candidate from Global Coarse Search using $\mathcal{L}_{\text{NC}}$
    \For{$\sigma = \sigma_{\text{max}}$ down to $\sigma_{\text{min}}$} \Comment{Dynamic Gaussian smoothing}
        \State $\mathbf{x}_{s,i} \leftarrow \mathbf{x}_{s,i} - \eta \cdot \nabla_{\mathbf{x}_{s,i}} \mathcal{L}_{\text{NC}} \left(\mathcal{F}(\hat{\mathbf{P}}_c, \mathbf{x}_{s,i}), \mathbf{S}_{B,i}; \sigma \right)$
    \EndFor
    \State $\tilde{\mathbf{X}}_B \leftarrow \tilde{\mathbf{X}}_B \cup \{\mathbf{x}_{s,i}\}$
\EndFor

\Statex
\State \textbf{ Stage 3: Robust Rigid Array Estimation}
\State $(\mathbf{R}_{\text{init}}, \mathbf{t}_{\text{init}}, \mathcal{I}) \leftarrow \text{Modified RANSAC}(\tilde{\mathbf{X}}_B, \mathbf{X}_{\text{template}})$ 
\State \Comment{$\mathcal{I}$ is the robust inlier subset passing geometric consistency pre-check}

\Statex
\State \textbf{ Stage 4: Global Differentiable Fine-tuning}
\State Initialize Euler angles $\bm{\theta}$ from $\mathbf{R}_{\text{init}}$, and translation $\mathbf{t} \leftarrow \mathbf{t}_{\text{init}}$
\While{not converged}
    \State $\mathcal{L}_{\text{global}} \leftarrow \sum_{i \in \mathcal{I}} \mathcal{L}_{\text{NC}} \left( \mathbf{R}(\bm{\theta})\mathbf{x}_{\text{template},i} + \mathbf{t} \right)$ \Comment{Inlier-masked loss}
    \State $\bm{\theta} \leftarrow \bm{\theta} - \alpha \cdot \nabla_{\bm{\theta}} \mathcal{L}_{\text{global}}$ \Comment{Update global rotation}
    \State $\mathbf{t} \leftarrow \mathbf{t} - \alpha \cdot \nabla_{\mathbf{t}} \mathcal{L}_{\text{global}}$ \Comment{Update global translation}
\EndWhile
\State $\mathbf{X}_B^{\text{fine}} \leftarrow \mathbf{R}(\bm{\theta})\mathbf{X}_{\text{template}} + \mathbf{t}$ \Comment{Kinematic constraint applied to ALL sensors}

\Statex
\State \textbf{ Stage 5: Joint Multi-view Reconstruction}
\State $\mathbf{X}_{\text{joint}} \leftarrow \mathbf{X}_A \cup \mathbf{X}_B^{\text{fine}}$
\State $\mathbf{S}_{\text{joint}} \leftarrow \mathbf{S}_A \cup \mathbf{S}_B$
\State $\mathbf{P}_{\text{final}} \leftarrow \arg\min_{\mathbf{P}} \left( \| \mathcal{F}(\mathbf{P}, \mathbf{X}_{\text{joint}}) - \mathbf{S}_{\text{joint}} \|_2^2 + \lambda \mathcal{L}_{\text{TGV}}(\mathbf{P}) \right)$

\State \Return $\mathbf{P}_{\text{final}}$, $\mathbf{X}_B^{\text{fine}}$
\end{algorithmic}
\end{algorithm}

\section{Results}\label{sec3}

\subsection{3D PA image reconstruction under multi-view sparse hemispherical array}\label{subsec3_1}

\begin{figure}[htbp]
\centering
\includegraphics[width=0.8\textwidth]{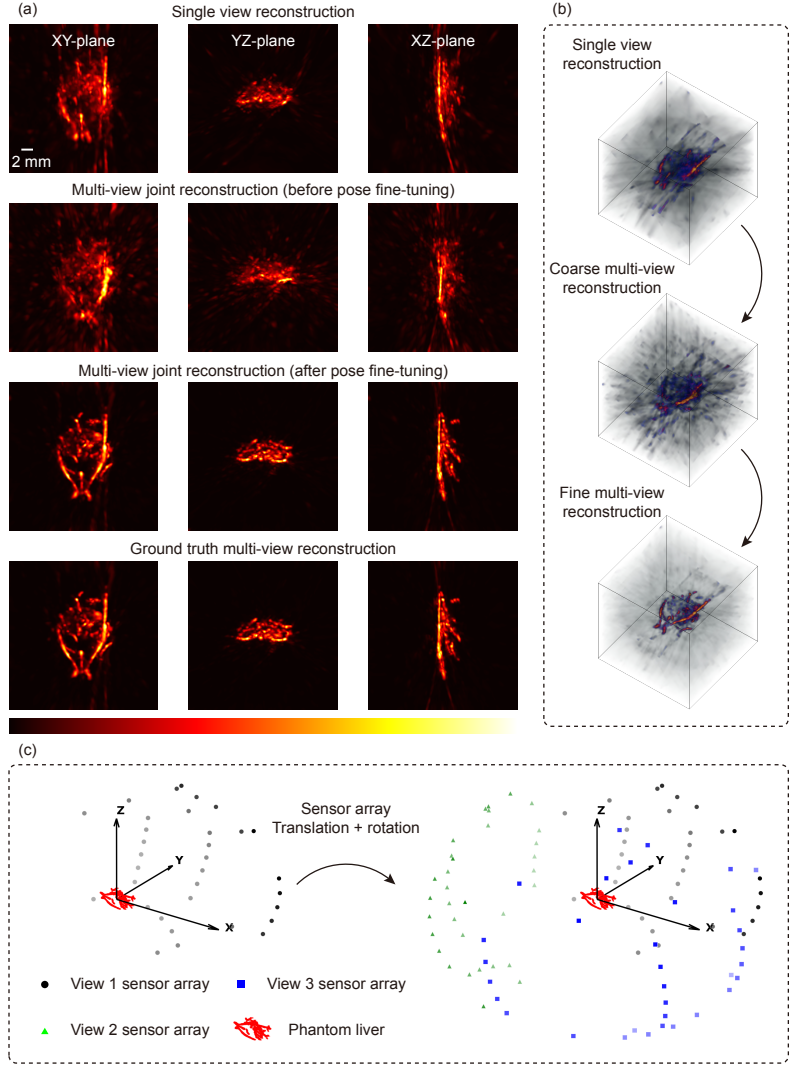}
\caption{The reconstruction results of multi-view simulation data with PA-SFM algorithm. (a) The single view reconstruction, multi-view coarse pose reconstruction, multi-view fine-tuned pose reconstruction and ground truth of the phantom liver (Scale: 2 mm). (b) The 3D results of single view, coarse multi-view and fine multi-view reconstruction. (c) The multi-view setup of the simulation experiment.}\label{simulation}
\end{figure}

\begin{figure}[htbp]
\centering
\includegraphics[width=1.0\textwidth]{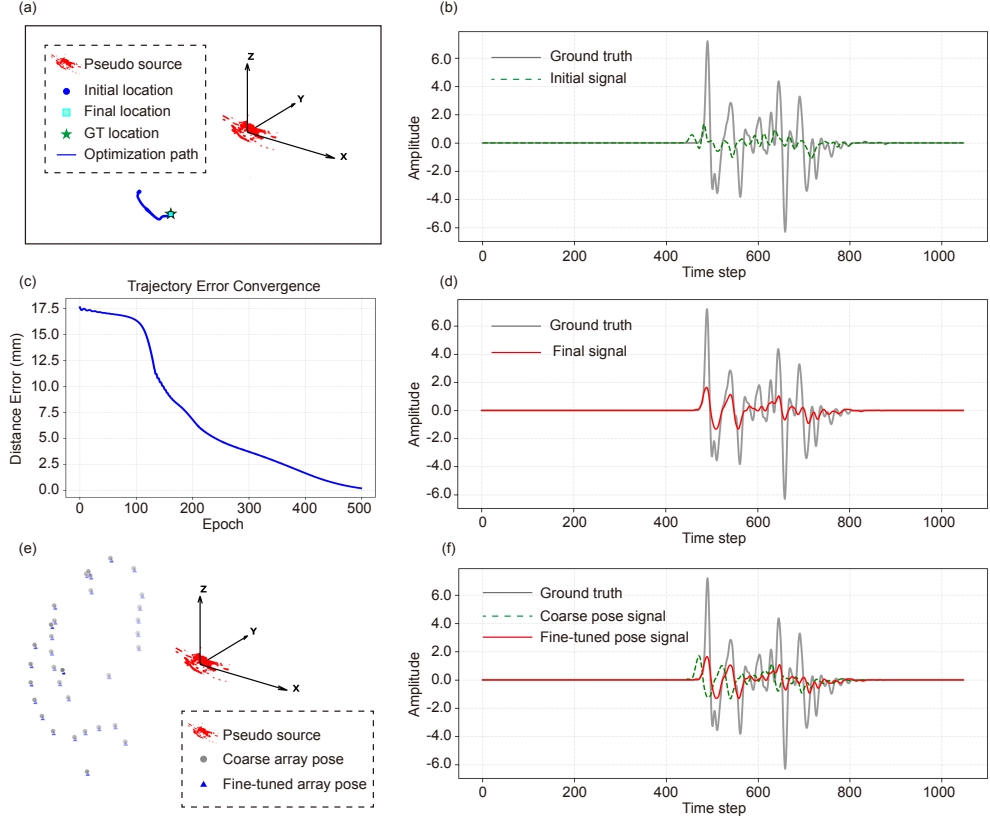}
\caption{The PA signal of the target sensor at different stages of the PA-SFM algorithm. (a) Differentiable single
sensor location optimization. (b) The PA signal of the target sensor at the initial location. (c) The distance error curve of the single sensor location optimization. (d) The PA signal of the target sensor at the final location. (e) Global differentiable pose fine-tuning of the new pose. (f) (b) The PA signal of the target sensor before and after the global differentiable pose fine-tuning.}\label{simu_sig}
\end{figure}

To evaluate the performance of the proposed PA-SFM algorithm, we first conducted numerical simulations using a phantom liver model under a multi-view setup. As shown in Fig.~\ref{simulation}(c), the simulation utilized a multi-view configuration to acquire sparse hemispherical array data. Specifically, the array consisted of a 33-element sparse planar array distributed across a spherical cap. View 1 served as the reference array pose, while View 2 and View 3 represented novel array poses generated by translating and rotating around the simulated source. To demonstrate the pose optimization mechanism during the simulation, Fig.~\ref{simu_sig} illustrates the PA signals of the target sensor at different optimization stages. The differentiable single-sensor localization effectively optimizes the sensor location, which is evidenced by a clear convergence in the distance error curve (Fig.~\ref{simu_sig}(c)). After filtering out aberrant single-sensor localizations via rigid constraints, a minor positional deviation persisted in the global pose. Consequently, the PA signal extracted from the coarsely predicted sensor pose exhibited a noticeable discrepancy compared to the actual ground-truth supervision signal (as indicated by the green dashed line for the coarse prediction and the solid gray line for the ground truth in Fig.~\ref{simu_sig}(f)). At this coarse estimation stage, the maximum and minimum Euclidean distance errors for sensor localization were recorded as $0.002334$~mm and $0.000265$~mm, respectively. Through further Global Differentiable Fine-tuning, the alignment of the PA signals improves progressively; the fine-tuned pose signal closely matches the ground-truth signal compared to both the initial and coarse pose stages (Fig.~\ref{simu_sig}(b-f)). Ultimately, the maximum and minimum sensor localization errors were substantially reduced to $0.001784$~mm and $0.000135$~mm, respectively, demonstrating the high accuracy of the proposed fine-tuning mechanism.

Based on this precise pose recovery, we systematically compared the 3D reconstruction results. As depicted in Fig.~\ref{simulation}(a-b), the single-view reconstruction suffers from severe limited-view artifacts, yielding quantitative metrics of $25.51$~dB for the peak signal-to-noise ratio (PSNR) and $0.6775$ for the structural similarity index measure (SSIM) in the Maximum Amplitude Projection (MAP) along the XY plane. While the multi-view joint reconstruction using the coarse pose estimation effectively expands the field of view, it still exhibits structural blurring due to misalignment, resulting in a degraded PSNR of $21.35$~dB and an SSIM of $0.4552$. However, after applying the PA-SFM global differentiable pose fine-tuning module, the refined multi-view reconstruction demonstrates a substantial improvement. It successfully mitigates motion artifacts and restores high-resolution 3D structures that are highly comparable to the ground truth of the phantom liver, achieving a significantly improved PSNR of $38.90$~dB and an SSIM of $0.9637$.

\subsection{In-vivo rat results of multi-view PA-SFM joint reconstruction}\label{subsec3_2}

\begin{figure}[htbp]
\centering
\includegraphics[width=0.9\textwidth]{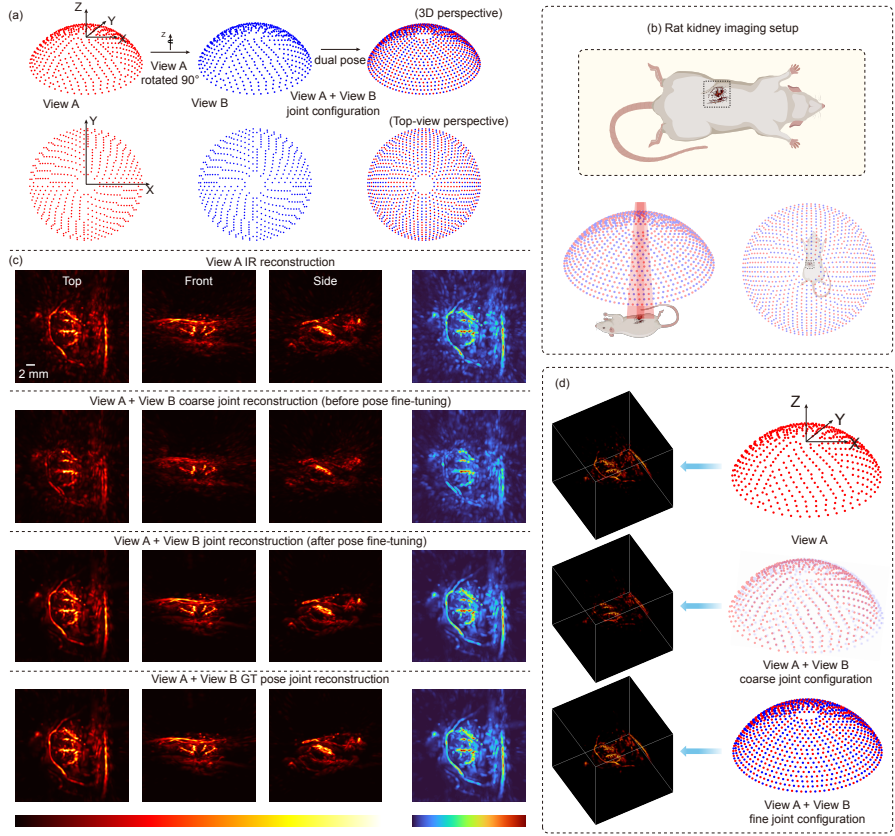}
\caption{The reconstruction results of multi-view \textit{in-vivo} data with PA-SFM algorithm. (a) The freehand multi-view 3D PAI for \textit{in-vivo} rat kidney imaging. (b) The rat kidney imaging setup. (c) The single view reconstruction, multi-view coarse pose reconstruction, multi-view fine-tuned pose reconstruction and ground truth of the \textit{in-vivo} rat kidney (Scale: 2 mm). (d)  The 3D results of single view, coarse multi-view and fine multi-view reconstruction.}\label{in_vivo}
\end{figure}

We further validated the clinical feasibility of the PA-SFM framework through \textit{in-vivo} multi-view 3D photoacoustic imaging of a rat kidney (Fig.~\ref{in_vivo}(a-b)), which was acquired by Professor Kim's lab using a hemispherical ultrasound (US) transducer array with 1024 elements and a radius of 60 mm~\cite{choi2023deep}. More details regarding the 3D PAI system and animal experiments can be found in literatures~\cite{choi2023recent, kim20243d, kim2022deep}. Specifically, to emulate a multi-view acquisition scheme under limited-view conditions, we deliberately partitioned the full 1024-element spherical array into two interleaved, 512-element sparse subsets. These two distinct subsets, designated as View A and View B, represent two independent imaging poses separated by a $90^\circ$ relative rotation along the azimuthal axis. The final \textit{in-vivo} 3D reconstruction results are presented in Fig.~\ref{in_vivo}. 

For a comprehensive evaluation, we compared the single-view Iterative Reconstruction against multi-view joint reconstructions utilizing the coarsely predicted pose, our PA-SFM fine-tuned pose, and the ground-truth (GT) pose. As detailed in Fig.~\ref{in_vivo}(c), the respective rows display the MAPs of the reconstructed volumes from the top (above), side, and front viewpoints utilizing a hot colormap. Furthermore, a top-view MAP utilizing a turbo colormap is explicitly provided to better visualize background noise and reconstruction artifacts. 

As shown in Fig.~\ref{in_vivo}(c) and \ref{in_vivo}(d), single-view methods provide limited structural information with prominent artifacts. In contrast, the multi-view fine-tuned pose reconstruction successfully fuses the dual-pose data (View A + View B), effectively overcoming limited-view constraints. Consequently, the PA-SFM framework yields a highly accurate, artifact-free 3D representation of the \textit{in-vivo} rat kidney vascular network, demonstrating high fidelity and structural integrity. 

Quantitative assessments based on the top-view MAP further corroborate these visual improvements. When evaluated against the GT multi-view reconstruction, the single-view approach yielded a PSNR of $30.20$~dB and a SSIM of $0.7019$. Similar to the simulation phase, the uncalibrated coarse-pose multi-view reconstruction suffered from severe misalignment blurring, leading to degraded metrics (PSNR: $23.58$~dB, SSIM: $0.6098$). Remarkably, after applying the PA-SFM fine-tuning module, the quantitative performance surged to a PSNR of $41.42$~dB and an SSIM of $0.9864$, confirming its exceptional capability in mitigating motion artifacts and recovering complex biological structures.

\section{Discussion}\label{sec4}

In this work, we propose the PA-SFM framework, a novel tracker-free approach that leverages exclusively single-modality photoacoustic data for simultaneous sensor pose recovery and high-fidelity 3D reconstruction. Traditional handheld 3D photoacoustic tomography systems typically rely on bulky and expensive external tracking hardware, such as optical tracking systems, electromagnetic trackers, or inertial measurement units, to correct manual motion artifacts. However, these conventional methods often suffer from strict line-of-sight requirements, susceptibility to electromagnetic interference, or inherent cumulative drift, which significantly limit their clinical accessibility and flexibility.
To overcome these hardware constraints, PA-SFM integrates the forward acoustic wave equation directly into a differentiable programming pipeline. Unlike conventional SFM methods based on visual features or traditional cross-correlation algorithms that fail in the inherently speckle-free photoacoustic domain, our method is built upon a physical understanding of acoustic propagation. By utilizing a high-performance, GPU-accelerated acoustic radiation kernel, the framework simultaneously optimizes the 3D photoacoustic source distribution and the 6-DoF sensor array pose via gradient descent.
Furthermore, ensuring robust convergence in unconstrained freehand scenarios poses a significant challenge. To address this, we implemented a comprehensive coarse-to-fine optimization strategy. By incorporating a modified RANSAC algorithm with geometric consistency pre-checks and rigid-body kinematics, the system effectively identifies and eliminates localization outliers. The global differentiable fine-tuning module utilizes the highly reliable signals from an inlier subset to physically guide unreliable sensors into their correct anatomical positions, thus guaranteeing high spatial precision.
Our validations, including both numerical simulations and \textit{in-vivo} rat kidney experiments, demonstrate that PA-SFM successfully achieves sub-millimeter positioning accuracy. By coherently fusing multi-view data to augment the synthetic aperture, the framework mitigates limited-view artifacts and restores high-resolution 3D vascular structures that are comparable to ground-truth benchmarks. Ultimately, PA-SFM provides a low-cost, software-defined solution that significantly improves the viability of routine clinical freehand photoacoustic imaging without the need for complex external positioning sensors.

\backmatter

\section*{Acknowledgements}

This research was supported by the following grants: the National Key R\&D Program of China (No. 2023YFC2411700, No. 2017YFE0104200); the Beijing Natural Science Foundation (No. 7232177); the National Natural Science Foundation of China (62441204, 62472213); the Basic Science Research Program through the National Research Foundation of Korea (NRF) funded by the Ministry of Education (2020R1A6A1A03047902).

\section*{Declarations}

\subsection*{Competing interests}

All authors declare no competing interests.

\subsection*{Data availability}

The data supporting the findings of this study are provided within the Article and its Supplementary Information. The raw and analysed datasets generated during the study are available for research purposes from the corresponding authors on reasonable request.

\subsection*{Code availability}

The source code for SlingBAG, along with supplementary materials and demonstration videos, has been made publicly available in the following GitHub repository: \href{https://github.com/JaegerCQ/PA-SFM}{https://github.com/JaegerCQ/PA-SFM}.

\subsection*{Author contributions}

Shuang Li and Jian Gao conceived and designed the study. Shuang Li and Jian Gao contributed to the design of algorithms. Chulhong Kim and Seongwook Choi provided the \textit{in-vivo} experimental data. Qian Chen, Yibing Wang, Shuang Wu and Yu Zhang contributed to the simulation experiments. Tingting Huang, Yucheng Zhou and Boxin Yao contributed to the metrics calculation. Changhui Li and Yao Yao supervised the study. All of the authors contributed to writing the paper.

\appendix

\section{Analytical Expression for Spherically Symmetric Photoacoustic Sources: theoretical foundation for the differentiable acoustic radiation}
\label{appendix:theoretical_derivation}

In this appendix, we present a comprehensive analytical derivation of the spatiotemporal acoustic pressure generated by photoacoustic sources, which serves as the theoretical foundation for the differentiable acoustic radiation engine proposed in Section \ref{sec2}. We start from the fundamental photoacoustic wave equation, derive closed-form solutions for spherically symmetric distributions (including the Gaussian model used in our framework), and systematically formulate the far-field approximations that enable the ultrafast GPU-accelerated forward simulation.

\subsection{Theoretical Background}
\subsubsection{Photoacoustic Wave Equation}
The pressure wave generated by the photoacoustic effect satisfies the following wave equation~\cite{xu2005universal,xu2006photoacoustic}:
\begin{equation}
\left(\nabla^2 - \frac{1}{v_s^2}\frac{\partial^2}{\partial t^2}\right)p(\vect{r},t) = -\frac{\beta}{C_p}\frac{\partial H(\vect{r},t)}{\partial t}
\label{eq:wave_eq}
\end{equation}
where $p(\vect{r},t)$ is the acoustic pressure, $v_s$ is the speed of sound, $\beta$ is the thermal expansion coefficient, $C_p$ is the specific heat capacity, and $H(\vect{r},t)$ is the heating function. For instantaneous heating, the initial conditions are:
\begin{equation}
p(\vect{r},0) = p_0(\vect{r}), \quad \frac{\partial p}{\partial t}(\vect{r},0) = 0
\label{eq:initial_cond}
\end{equation}

\subsubsection{Fundamental Integral Formula}
The solution to the photoacoustic wave equation can be expressed using Green's function as~\cite{xu2005universal,xu2006photoacoustic}:
\begin{equation}
p(\vect{r},t) = \frac{1}{4\pi v_s^2}\frac{\partial}{\partial t}\left[\frac{1}{v_s t}\int d\vect{r}'\,p_0(\vect{r}')\,\delta\!\left(t-\frac{\abs{\vect{r}-\vect{r}'}}{v_s}\right)\right]
\label{eq:integral_form}
\end{equation}
where $p_0(\vect{r}')$ is the initial pressure distribution.

\subsection{Derivation for Spherically Symmetric Case}
Assume the initial pressure distribution is spherically symmetric about the origin, $p_0(\vect{r}') = p_0(r')$, where $r' = \abs{\vect{r}'}$ and $r = \abs{\vect{r}}$. Using the scaling property of the delta function, $\delta(t-\abs{\vect{r}-\vect{r}'}/v_s) = v_s\,\delta(\abs{\vect{r}-\vect{r}'}-v_s t)$, and substituting it into Eq.~\ref{eq:integral_form}, we get:
\begin{equation}
p(\vect{r},t) = \frac{1}{4\pi v_s^2}\frac{\partial}{\partial t}\left[\frac{1}{t}\int d\vect{r}'\,p_0(r')\,\delta\!\left(\abs{\vect{r}-\vect{r}'}-v_s t\right)\right]
\label{eq:pressure_intermediate}
\end{equation}

By transforming to spherical coordinates $(r',\theta,\phi)$ and applying the law of cosines $\abs{\vect{r}-\vect{r}'} = \sqrt{r^2 + r'^2 - 2rr'\cos\theta}$, the delta function can be transformed (see Section \ref{sec:detailed_derivations}) as:
\begin{equation}
\delta\!\left(\abs{\vect{r}-\vect{r}'}-v_s t\right) = \frac{v_s t}{rr'}\,\delta\!\left(\cos\theta - \frac{r^2 + r'^2 - (v_s t)^2}{2rr'}\right)
\end{equation}

Integrating over solid angles yields $\int d\Omega\,\delta(\dots) = \frac{2\pi v_s t}{rr'}$ under the condition $\abs{r-v_s t} \leq r' \leq r+v_s t$. Substituting this back into the pressure expression and simplifying yields:
\begin{equation}
p(r,t) = \frac{1}{2v_s r}\frac{\partial}{\partial t}\left[\int_{\abs{r-v_s t}}^{r+v_s t} r'p_0(r')\,dr'\right]
\label{eq:pressure_simplified}
\end{equation}

Applying Leibniz's rule for differentiating under the integral sign gives the unified analytical expression for the spherically symmetric case:
\begin{equation}
\boxed{p(r,t) = \frac{1}{2r}\left[(r+v_s t)p_0(r+v_s t) + (r-v_s t)p_0(\abs{r-v_s t})\right]}
\label{eq:final_expression}
\end{equation}

\subsection{Analytical Solutions for Typical Initial Distributions}
Based on Eq.~\ref{eq:final_expression}, explicit analytical solutions can be derived for several commonly encountered distributions:

\textbf{1. Uniform Spherical Source ($p_0(r) = p_0\,U(a_0 - r)$):}
For an observation point outside the sphere ($r > a_0$), $p(r,t) = \frac{p_0}{2r}(r - v_s t)$ when $r - a_0 \leq v_s t \leq r + a_0$, and $0$ otherwise.

\textbf{2. Gaussian Distribution ($p_0(r) = p_c\,\exp(-r^2/2\sigma^2)$):}
\begin{equation}
p(r,t) = \frac{p_c}{2r}\left[(r+v_s t)e^{-\frac{(r+v_s t)^2}{2\sigma^2}} + (r-v_s t)e^{-\frac{(r-v_s t)^2}{2\sigma^2}}\right]
\label{eq:gaussian_solution}
\end{equation}

\textbf{3. Exponential Distribution ($p_0(r) = p_c\,e^{-r/a}$):}
\begin{equation}
p(r,t) = \frac{p_c}{2r}\left[(r+v_s t)e^{-(r+v_s t)/a} + (r-v_s t)e^{-\abs{r-v_s t}/a}\right]
\label{eq:exponential_solution}
\end{equation}

\textbf{4. Power-Law Distribution ($p_0(r) = A/(r^2 + a^2)^{\nu}, \nu > 1/2$):}
\begin{equation}
p(r,t) = \frac{A}{2r}\left[\frac{r+v_s t}{\big((r+v_s t)^2 + a^2\big)^{\nu}} + \frac{r-v_s t}{\big((r-v_s t)^2 + a^2\big)^{\nu}}\right]
\label{eq:power_law_solution}
\end{equation}

\subsection{Far-Field Approximations}
Under far-field conditions ($r \gg \text{characteristic source dimension}$ and $t \approx r/v_s$), the first term in Eq.~\ref{eq:final_expression} (representing an inward converging wave) is exponentially or power-law suppressed. The dominant contribution comes from the second term, leading to the unified far-field approximation:
\begin{equation}
p(r,t) \approx \frac{1}{2r}(r-v_s t)p_0(\abs{r-v_s t})
\label{eq:far_field_general}
\end{equation}

Applying this to the \textbf{Gaussian distribution} yields the outward propagating Gaussian pulse:
\begin{equation}
p(r,t) \approx \frac{p_c}{2r}(r-v_s t)\exp\!\left(-\frac{(r-v_s t)^2}{2\sigma^2}\right)
\label{eq:gaussian_far_field}
\end{equation}
which provides the direct mathematical justification for the Gaussian kernel $\mathcal{G}$ defined in Eq.~1 of the main text.

\subsection{GPU-Accelerated Forward Simulation Framework}
Although the analytical expressions provide closed-form solutions, practical PA-SfM simulations involve a very large number of discrete source elements ($N_s$) and detector locations ($N_d$). To address this, we discretize the far-field approximation (Eq.~\ref{eq:far_field_general}) to build the GPU-accelerated forward model. The acoustic pressure measured at a detector position $\mathbf{r}_d$ becomes a superposition of outgoing waves emitted by all source elements $\mathbf{r}_k$ with amplitudes $P_c^{(k)}$:
\begin{equation}
p(\mathbf{r}_d,t) \approx \sum_{k=1}^{N_s} \frac{1}{2 r_k} \left( r_k - v_s t \right) P_c^{(k)} \, \mathcal{K}\!\left(\abs{r_k - v_s t}\right),
\label{eq:gpu_forward_model}
\end{equation}
where $r_k = \lVert \mathbf{r}_d - \mathbf{r}_k \rVert$ and $\mathcal{K}(\cdot)$ denotes the designated kernel function. Based on Eq.~\ref{eq:gaussian_far_field}, the Gaussian kernel is defined as $\mathcal{K}_{\mathrm{G}}(\Delta r) = \exp(-\Delta r^2 / 2 \sigma^2)$.

This discrete model exhibits two independent dimensions of parallelism: the detector index and the temporal sampling index. In our PyTorch-Triton implementation, a two-dimensional GPU execution grid is employed. Distances $r_k$ and time delays are computed on-the-fly, avoiding the explicit construction of large distance matrices. This design enables the efficient evaluation of $\mathcal{O}(N_s \times N_d \times N_t)$ operations, seamlessly supporting bidirectional gradient flow for the PA-SfM framework. Open-source implementations for this generalized ultrafast forward simulation are also publicly available at \url{https://github.com/JaegerCQ/PA-SFM}.

\subsection{Detailed Delta Function Derivations}
\label{sec:detailed_derivations}
Starting from $\delta(\sqrt{A} - R)$ where $A = r^2 + r'^2 - 2rr'\cos\theta$ and $R=v_st$. Let $g(A) = \sqrt{A} - R$, then $\frac{dg}{dA} = \frac{1}{2\sqrt{A}}$. At $A = R^2$, $\abs{\frac{dg}{dA}} = \frac{1}{2R}$. Using the delta function transformation formula:
\begin{equation}
\delta(g(A)) = \frac{\delta(A - R^2)}{\abs{g'(R^2)}} = 2R\,\delta(A - R^2)
\end{equation}
Next, let $h(\cos\theta) = r^2 + r'^2 - R^2 - 2rr'\cos\theta$. The zero point is $\cos\theta_0 = \frac{r^2 + r'^2 - R^2}{2rr'}$ and the derivative is $\abs{\frac{dh}{d(\cos\theta)}} = 2rr'$. Therefore:
\begin{equation}
\delta(h(\cos\theta)) = \frac{1}{2rr'}\,\delta\!\left(\cos\theta - \frac{r^2 + r'^2 - R^2}{2rr'}\right)
\end{equation}
Combining these factors, we obtain:
\begin{equation}
\delta\!\left(\abs{\vect{r}-\vect{r}'} - R\right) = \frac{R}{rr'}\,\delta\!\left(\cos\theta - \frac{r^2 + r'^2 - R^2}{2rr'}\right)
\end{equation}

\subsection{Nomenclature}
\begin{table}[h]
\centering
\begin{tabular}{ll}
\toprule
\textbf{Symbol} & \textbf{Description} \\
\midrule
    $p(\vect{r},t)$ & Acoustic pressure at position $\vect{r}$ and time $t$ \\
    $p_0(\vect{r})$ & Initial pressure distribution \\
    $v_s$ & Speed of sound \\
    $\beta$ & Thermal expansion coefficient \\
    $C_p$ & Specific heat capacity \\
    $r$ & Radial distance from origin \\
    $t$ & Time \\
    $\delta(x)$ & Dirac delta function \\
    $\sigma$ & Gaussian width parameter \\
    $a_0$ & Radius of uniform sphere \\
\bottomrule
\end{tabular}
\end{table}

Please refer to our separate Supplementary Information file for further details.

\bibliography{references}% common bib file

\end{document}